\documentclass[runningheads]{llncs}

\usepackage[T1]{fontenc}
\usepackage{graphicx}
\usepackage{amsmath}
\usepackage{amsfonts}
\usepackage{multirow}
\usepackage{array}
\newcolumntype{L}[1]{>{\raggedright\arraybackslash}p{#1}}
\usepackage{makecell}
\usepackage{tabularx}
\usepackage{booktabs}
\usepackage{xparse}
\usepackage{xcolor}
\usepackage{color,soul}
\usepackage{dashrule}
\usepackage{tikz}
\usepackage{multicol}
\usepackage{url}
\usepackage[hidelinks]{hyperref}
\usepackage{subcaption}

\newcounter{algorithm}

\NewDocumentCommand{\rot}{O{45} O{1em} m}{\makebox[#2][l]{\rotatebox{#1}{#3}}}

\title{PVRA: A Pointwise Key-point Voting Framework for Robotic Assembly}
\titlerunning{PVRA: A Pointwise Key-point Voting Framework for Robotic Assembly}

\author{Kulunu Samarawickrama, Roel Pieters}
\authorrunning{Kulunu et al.}
\institute{Tampere University, Finland}

\begin{document}
\maketitle

\begin{abstract}

 Modern computer vision has enabled partial autonomy in robotic assembly manipulation. However, performing autonomous manipulation of a progressive assembly demands a more specific set of skills, in addition to perceiving the objects.  Through a comparative analysis of research in the associated domains, we deduce that object-centric perception must advance towards learning assembly dependencies to predict meaningful actionable outputs for autonomous assembly manipulation. Subsequently, we present a 3D keypoint-based modular learning framework to learn assembly dependencies to infer actionable outputs given a RGB-D input of an assembly scene. We train and evaluate our trained network on an assembly pose estimation dataset and compare it against object-centric baselines with an augmented set of metrics for progressive assemblies. The source code of the proposed framework is hosted at \textit{\url{https://github.com/KulunuOS/PVRA}}

\end{abstract}

\section{Introduction}\label{sec:intro}

Assembling parts together to obtain a final product is a trivial task for humans. However, achieving a similar level skill for a robot is a complex challenge to date. Research on robotic assembly spans over multiple domains with diverse applications and variable constraints. Some examples include; robotic manufacturing, service robots, medical robots and applications, space explorations, fractured part reconstruction in archaeology, repair and restoration applications, etc. There is continuous demand for advanced assembly capabilities in robots despite extensive amount of studies. Although current research addresses these challenges via visual perception and reinforced learning, it remains debatable whether these exhibit behaviors of an intelligent robot with assembly task awareness. In this paper we introduce our notion of assembly task awareness and an end-to-end learning method to enable it for robotic manipulation.

Computer vision tools such as object detection, classification and semantic segmentation are paramount to endow perception to robots. while these tools provide basic scene understanding, robotic manipulation  requires 3D spatial understanding of the robot environment. Consequently, 6 DoF object pose estimation emerged as a critical component in robotic perception which has become pivotal in robotic grasping and object placement, over the time. However, a multi-object assembly task imposes more requirements on context awareness. Spatial awareness is crucial in localizing pre-assembled and post-assembled poses of assembly objects. Assembling the objects in an optimal sequence requires temporal awareness. Inter-part relational awareness decides the structural coherence of a final assembly. Moreover, awareness on the stability under gravity is an essential physical feasibility.

The above observations formulate the main components of task awareness required for a progressive assembly task. This awareness can be interpreted as the ability of an entity to perceive an assembly configuration and reason about its spatial, temporal and relational dependencies. A module that can learn to produce assembly-aware actionable outputs is extremely beneficial for robotic assembly manipulation. This is critical for modern robotics as it enables manipiulation capabilities beyond primitive pick and placement operations towards task-aware manipulation. To this end we introduce our method PVRA: A Pointwise Key-point Voting Framework for Robotic Assembly. The main contributions of this paper are as below:
\vspace{-0.6em}
\begin{enumerate}
    \setlength{\itemsep}{0.1em}
    \setlength{\topsep}{0pt}
    \setlength{\parsep}{0pt}
    \renewcommand{\labelenumi}{\roman{enumi}.}
    \item A 3D keypoint-based modular learning framework to enable assembly task awareness for robotic entities
    \item An augmented set of metrics to benchmark assembly-ware role segmentation and pose estimation.
    \item Implementation and evaluation of the proposed framework on a custom object assembly dataset
    \item A comparative evaluation of the proposed model against object-centric baselines. 
\end{enumerate}

\section{Related Work}\label{sec:related_work}

Assembly has long been a major research topic in process engineering, particularly under the domains of Assembly/\allowbreak Disassembly Sequence Planning (ASP/DASP) and Assembly/\allowbreak Disassembly Path Planning (APP/DAPP). The main aim of such work is to optimize manufacturing processes by Assembly/Disassembly planning (AP/DAP)
\cite{Ghandi2015}, \cite{Wilson1994}. Traditionally, these focused on deterministic planning of actions and trajectories based on geometric modeling and combinatorial reasoning. However, with the rapid advancements in robotics and automation, a research direction focusing on transforming classical AP to a cognitive skill based on dynamic field of artificial intelligence has emerged. While this shares the same goals as AP, the majority of existing research focused on enabling AP via perception-driven and learning-based methods. The Table \ref{tab:literature} summarizes state-of-the-art methods and their respective learning concepts, network architectures, inputs, outputs and datasets utilized.

\clearpage
\begin{center}
\refstepcounter{table}\label{tab:literature}
{\small Table~\thetable: Categorization of assembly pose estimation methods. GNN denotes Graph Neural Network, VNN denotes Vector Neuron Network, CNN denotes Convolutional Neural Network.\par}
\vspace{2mm}
\renewcommand{\arraystretch}{1.08}
\setlength{\tabcolsep}{3pt}
\footnotesize
\begin{tabular}{|L{0.20\textwidth}|L{0.22\textwidth}|L{0.48\textwidth}|}
\hline
\textbf{Method (Year)} & \textbf{Learning Concept} & \textbf{Architecture / Input / Output / Dataset} \\
\hline
ASAP \cite{Tian2023} (2022) &
Assembly by disassembly &
GNNs; Point cloud; Assembly sequence; Custom \cite{Tian2023} \\
\hline
GPAT \cite{GPAT} (2023) &
Target segmentation &
Transformers; Point cloud; Assembly pose; PartNet \cite{partnet} \\
\hline
PAST \cite{Zhu2023} (2023) &
Assembly by disassembly &
Transformers; Point cloud; Assembly sequence and assembly pose; D4PAS \cite{Zhu2023} \\
\hline
Neural Shape Mating \cite{Chen2022} (2022) &
Shape priors reconstruction &
Adversarial Network; Point cloud; Assembly pose; Thingi10k \cite{thingi10k}, GSO \cite{GSO}, Shapenet \cite{shapenet2015} \\
\hline
RGL-Net \cite{Harish2022} (2022) &
Progressive relational reasoning &
Recurrent GNN; Point cloud; Assembly pose; PartNet \cite{partnet} \\
\hline
Instance Encoded Transformer \cite{Zhang2022} (2022) &
Inter-part relational learning &
Transformers; Point cloud; Assembly pose; PartNet \cite{partnet} \\
\hline
Dynamic Graph Learning \cite{Huang2020} (2020) &
Part relation reasoning &
GNN; Point cloud; Assembly pose; PartNet \cite{partnet} \\
\hline
SE-3 Part Assembly \cite{Wu2023} (2023) &
Part correlations via SE(3) equivariance &
VNN; Mesh files; Assembly pose; GSM dataset \cite{Chen2022}, Breaking-Bad \cite{breaking} \\
\hline
This work: PVRA (2025) &
Modular learning &
CNN; RGB-D; Assembly cognition; 6DAPose \cite{6DAPosedataset} \\
\hline
\end{tabular}
\end{center}

\subsection{Assembly Sequence Planning}
Assembly Sequence Planning (ASP) constitutes the task of inferring a physically feasible and collision-free sequence of operations on assembly objects leading to an assembled product. "Assemble them All" \cite{Assemblethemall} proposes a purely simulation based collision-free tree-search strategy for ASP.  ASAP \cite{Tian2023} builds upon the same concept and extends it to a learnable model that accounts for gravitational stability of components, which enhances its applicability to real-world applications. The core principle employed in these works to explore the physical feasibility is called "Assembly by Disassembly". Exploiting this principle, PAST \cite{Zhu2023} proposes a transformer network with reasoning capability for ASP at multiple levels of the assembly, based on point clouds and assembly pose estimation as an auxiliary task. 

\subsection{Assembly Pose Estimation}
Given a set of parts or components Assembly Pose Estimation (APE) entails inferring the target object's 6 DoF pose in the assembled product while satisfying the assembly constraints. APE, in the absence of any prior knowledge about semantics of the parts, can be defined as generalized part assembly. Generalized part assembly is capable of assembling novel or unseen objects given a target assembly. GPAT \cite{GPAT} addresses this problem through "Target segmentation", where the network explicitly segments the target blue print and infers the assembly poses. However, GPAT assumes a global and single-step APE, which overlooks any intermediate steps in a progressive assembly task. This creates an adverse effect on feasibility and accuracy in performing a progressive assembly task due to the lack of progressive reasoning and incorporating physical constraints. 

Progressive part assembly requires more specific reasoning capabilities such as temporal awareness and inter-part relationships. "Instance Encoded Transformer" \cite{Zhang2022} introduces an approach to implicitly learn a sequential model for progressive assembly using an attention mechanism and instance encoding. It is effective in APE scenarios where the assembly sequence is critical. However, this approach relies on assemblies with a fixed sequence, limiting its temporal awareness. Therefore, it is less effective in scenarios with assembly ambiguities.  RGL-Net \cite{Harish2022} proposes a recurrent graph-based learning method which explicitly learns inter-part relationships with sequence encoding.  It features iterative refinement of APE over time by updating part relationships in the graph structure, leading to improved generalization. Similar work by Huang et al. \cite{Huang2020} implements a dynamic graph learning method which learns to assemble parts from a coarse-to-fine manner. It is robust against assembly ambiguities with better temporal awareness, making it suitable for assemblies with variable part configurations and flexible sequences.

Apart from the methods which aim at modeling and learning inter-part relationships, there exist other modern methods which leverage geometric priors for APE. For example, "Neural shape mating" \cite{Chen2022} employs adversarial networks to learn how to mate two shapes by first bisecting them into two complementary parts. This enables APE in a self-supervised manner, using only geometric alignment without contextual knowledge on semantics or structure.  Another study presents how the SE(3)-Equivariance \cite{Wu2023} quality can be exploited to learn transformation-consistent geometric features using Vector Neuron Networks (VNNs). These features, which maintain a predictable behavior under 3D rotations and translations, can be aggregated to generate pose-invariant representations of parts, enabling multi-object APE in a one-shot fashion. However, they are not capable of learning temporal dependencies, inter-part relationships or assembly ambiguities, which are essential for progressive assembly tasks.

\subsection{Key-Point based methods}
Point cloud-based methods have proven to be more useful for robot perception with the advent of RGB-D sensors due to their capability in extracting 3D spatial data from the real world \cite{frustum}\cite{votenet}\cite{densefusion}. However, point clouds obtained from RGB-D sensors can observe only a partial surface and are often noisy. Moreover, learning directly on point clouds becomes complex due to the size of point clouds. Alternatively, keypoint-based learning methods overcome above drawbacks as they learn on a unique sub-sample of a large point cloud as shown by PVNet\cite{Pvnet} and PVN3D\cite{PVN3D}. These methods are robust against occlusions, making them ideal for practical robot applications.  Our proposed method is inspired by PVN3D \cite{PVN3D}, which solely aims on object localization. In contrast, we re-purpose the foundation for enabling novel assembly cognition, addressing broader challenges in the robotic assembly domain.

\subsection{Assembly Datasets}
As evident from Table \ref{tab:literature}, a majority of methods use the popular PartNet \cite{partnet} dataset to train and evaluate their approach. PartNet contains 24 classes that belong to general everyday categories with part-level annotations. There are ample datasets available for geometry-based APE that mainly contain mesh files of objects, such as Google Scanned Objects \cite{GSO}, Thingi10k \cite{thingi10k}, etc. However, these datasets lack crucial information required for progressive part assembly, such as the assembly sequence, intermediate assembly steps, stability under gravity, and other ambiguities. To address this gap, ASAP \cite{Tian2023} presents their own custom dataset with 2,146 assemblies, with a subset of 240 stable assemblies to train their assembly sequence planning method. Similarly, PAST \cite{Zhu2023} presents D4PAS, a dataset curated for progressive assembly tasks with 1500+ mechanical part assemblies. In this work, we use the 6DApose object assembly dataset \cite{6DAPosedataset}, which extends the standard object pose estimation benchmark dataset format; BOP \cite{BOP}.

\section{Problem Definition}\label{sec:problem}

Given an RGB-D input \( I_{RGBD} \) of an assembly scene with a set of assembly objects \( \{O_i\}_{i=1}^{C} \) in either disassembled or partially assembled state, where \( C \) is the total number of objects, the final assembled product \( H \) can be defined as 
where  \( O_i \) is the object to be assembled and its respective 6 DoF assembly pose \( T_i \in SE(3) \). Instead of predicting all poses in a one-shot manner, we formulate a progressive assembly approach by assembling parts sequentially over a fixed number of steps  \( t \in \{1, 2, \dots, C\} \). Therefore progressive assembly \( H_t \) at step \( t \) can be represented as 
    \begin{equation}
     H_t = H_{t-1} \;\cup\; T_t\bigl(O_{target}\bigr), \quad t = 1,2,\dots,C,
    \label{eq:0.2}
    \end{equation}
where \( O_{target} \) is the \textbf{target object} to be assembled at current step and its assembly pose is \( T_i \in SE(3) \). An example of a progressive object assembly consisting of 4 assembly steps is illustrated in Fig \ref{fig:framework}. Furthermore, we define a set of assumptions on progressive assemblies addressed in this work. The progressive assembly is assumed  to have (i) a fixed assembly sequence, (ii) a single and explicitly defined target object and its 6 DoF pose per assembly step and (iii) assembly objects are stably positioned under gravity. To further simplify the assembly complexity we assume, (iv) only one object is in contact with the target object when it's assembled, which is hereafter referred to as the \textbf{base object}. The reason for these assumptions is due to the complex nature and ambiguities in assembly tasks, which extends beyond the scope of the research problem, and to enable testability while maintaining reproducibility to implement our proposed method on different object assemblies. Thus, given an \( I_{RGB-D} \) dataset with relevant ground-truth data of a progressive assembly \( H_t \), the primary objective of our proposed method is to learn a supervised function \( f\bigl(I_{RGB-D}\bigr) \) that identifies the target object \( O_t \) and localizes its 6 DoF pose \( Q_t \in SE(3) \), and estimates its assembly pose \( T_t \in SE(3) \) w.r.t to camera optical frame. 


\section{Proposed Method}\label{sec:proposed_method}

\begin{figure}[hb!]
    \centering
    \includegraphics[width=\textwidth]{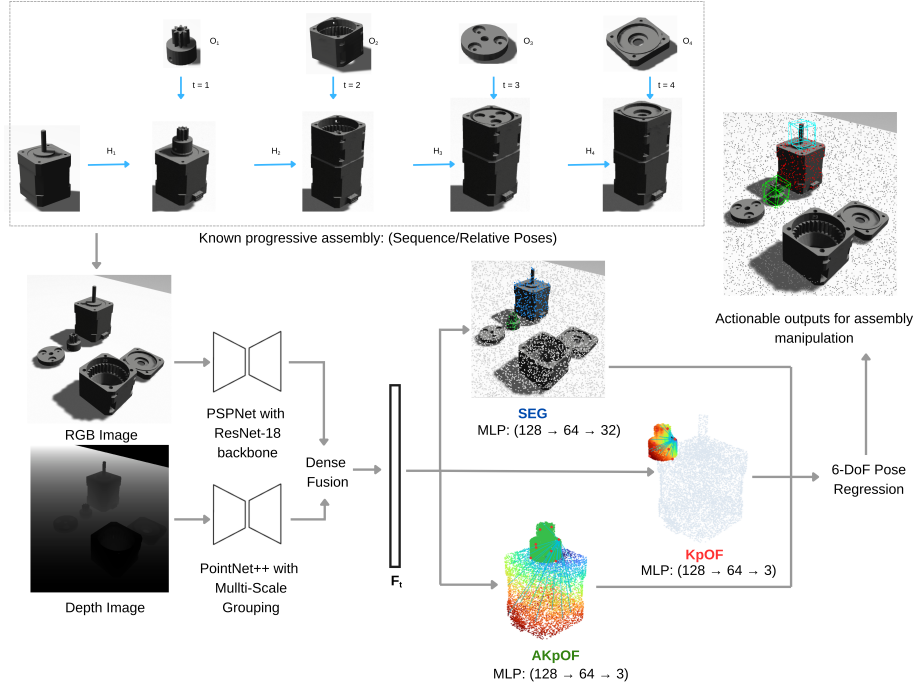}
    \caption{Proposed 3D keypoint-based learning framework for assembly cognition. MLP denotes Multi Layer Perceptrons.}
    \label{fig:framework}
\end{figure}

We propose PVRA: a supervised RGB-D point-cloud model (see Fig. \ref{fig:framework}) for progressive assembly under the specific assumptions described in Section \ref{sec:problem}. 
\clearpage
The method perceives a RGB-D input of the assembly scene and extracts features to learn a mapping that produces assembly task-aware actionable outputs; point-wise semantic roles ( target , base and background), 6-DoF pose of target object and its 6-DoF assembly pose. PVRA model does not represent a complete controller that executes an assembly task, but it predicts task-aware actionable outputs that enable such downstream tasks for autonomous agent to execute an assembly action at current step of a progressive assembly. The model shares RGB-D feature fusion strategy from PVN3D\cite{PVN3D}, but adopts prediction heads and supervision towards progressive assembly task.

\subsection{RGB-D Feature extraction and Fusion}

The depth image from the RGB-D input is back-projected into a point-cloud and sampled in to \( N\) points. Each point represents features: 3D coordinate, RGB value and surface normal. A pre-trained CNN-based network extracts semantic features from RGB input while the sampled pointcloud is processed by Pointnet++ feature extractor. The two feature streams are fused to a shared feature representation.
For \( N\) number of visible points sampled from a RGB-D input of \( H_t\),  the extracted feature map will be hereafter referred as \( F_t \). 

\subsection{Semantic Role Segmentation Module (SEG)}


The module predicts a per-point semantic probability map as a probability distribution \( S_t\) over \( C\) object classes: target, base and the background. In contrast to conventional semantic segmentation, these classes depend on the current assembly step and same object may be a target in one step and a base in a later step. The segmentation head outputs point-wise role logits. The learning model for module  \( {SEG}\) is described below where  \({S}_t(p) \in \mathbb{R}^C\) and \({S}_t^{(c)}(p)\) is the probability that point \( p\) belongs to object class \( C\). 
\begin{equation}S_t =  \mathrm{SEG}(F_t),
\quad
\sum_{c=1}^C{S}_t^{(c)}(p)=1.
\label{eq:eq_1}
\end{equation}



\subsection{Key-point offset prediction module (KpOF)}


The \({KpOF}\) module predicts a set of Manhattan distances \( L_{t}\) from a visible set of points \( \{p_i\}_{i=1}^{N} \) to a predefined set of 3D keypoints of the target object at pre-assembled state. 
This head is supervised only on the target-role points and the target 6-DOF pose is recovered by calculating the optimal transformation between predicted keypoints and object-model keypoints.
\begin{equation}
L_t =  \mathrm{KpOF}(F_t).
\label{eq:eq_3}
\end{equation}




\subsection{Assembly key-point offset prediction module (AKpOF)}
\({AKpOF}\) module predicts Manhattan distance offsets \( D_t \) from same set of visible points \( \{p_i\}_{i=1}^{N} \) to the same set of predefined 3D keypoints of the target object in assembled state. This head is supervised only on the base-role points and the target 6-DOF assembly pose is recovered by calculating the optimal transformation between predicted keypoints and object-model keypoints.

\begin{equation}
D_t =  \mathrm{AKpOF}(F_t).
\label{eq:eq_6}
\end{equation}


\subsection{Key-point Sampling}

Key-points are manually annotated for all assembly objects using their respective CAD files using farthest point sampling algorithm. They serve as ground-truth inputs for subsequent key-point offset prediction modules \({KpOF}\) and \({AKpOF}\). We initialize 8 key-points for each object as a pragmatic design choice which provides sufficient representation of spatial extent of objects and a manageable and consistent output dimensionality of key-point offset prediction heads (\( K \times 3 \) channels per object). The number of keypoints are fixed and object scale independent as the networks learn scale normalized distance offsets rather than absolute values. The aim is to sample sufficient sparse set of geometric anchors from which offsets can be inferred rather than densely representing the object. Although we do not present a full ablation over the number of keypoints, our work is consistent with prior 3D keypoint-based pose estimation work \cite{PVN3D}.

\subsection{Learning Objective}
PVRA shares a joint learning objective over the three shared prediction heads. In contrast to multi-task learning where each module learns an independent task, PVRA learns the same progressive assembly task. We jointly supervise the modules \({SEG}\), \({KpOF}\) and \({AKpOF}\)  using a combined loss function \(\mathcal{L}\). Segmentation head adopts focal loss and offset prediction heads adopt offset loss.
\begin{equation}
 \mathcal{L} = ( 2 \cdot \mathcal{L}_{\mathrm{SEG}} + \mathcal{L}_{\mathrm{KpOF}} + \mathcal{L}_{\mathrm{AKpOF}})/4. 
\label{eq:eq_10}
\end{equation}

\section{Evaluation}\label{sec:evaluation}

\subsection{Experimental setup}
We train and validate PVRA on the Nema17 gear-reducer progressive assembly dataset\cite{6DAPosedataset}. The dataset contains simulated assembly scenes with ground-truth object poses, assembly poses, RGB-D frames at each assembly step and CAD files of associated objects. The dataset consists of 431 assembly instances of 5 assembly objects and 4 assembly steps. This yields 8620 instances which are used to generate 60\% train split , 20\% validation split and a 20\% test split. The network was trained on a high-performance computing cluster consisting of 4 NVIDIA Tesla V100 GPUs with 128GB of VRAM. The source code is hosted at the GitHub repository: \url{https://github.com/KulunuOS/PVRA}.

A direct comparison of PVRA against related work summarized in Table \ref{tab:literature} is non-trivial. PVRA infers assembly task-aware actionable outputs on partial pointclouds obtained from RGB-D view samples grounded in a reference frame. In contrast most related methods expect complete pointclouds representing the assembly blueprints and they differ in problem formulations and assumptions. We compare PVRA against two CAD based pose estimation baselines. First, CAD-ICP-PCA, which initializes pointcloud registration from mask centroids and principal component axes before ICP refinement. Second, a modern CAD-based RGB-D pose estimator: FoundationPose\cite{foundationpose}. These methods estimate 6-DoF pose of target and base object and calculate assembly pose with known base-target transformation, whereas PVRA has to infer above outputs with learned assembly awareness. Both baselines, assume ideal object segmentation but PVRA predicts role labels on sampled RGB-D points. Therefore, we compare their performance under both GT-masks which represent dense object silhouettes and PVRA-predicted-masks with sparse points. This compares the performance of PVRA against baselines when object boundaries are under-represented in segmentation. 

\subsection{Metrics}

\noindent\textbf{Step-Segmentation Accuracy (SSA):}
  SSA measures assembly step-level segmentation accuracy using the average
  Intersection-over-Union (IoU) of the target and base roles. It reflects PVRA's
  ability to identify the active target and base objects with temporal and
  relational awareness:
  \begin{equation}
  \text{SSA}
  =
  \frac{1}{2}
  \left(
  \text{IoU}_{\text{target}}
  +
  \text{IoU}_{\text{base}}
  \right).
  \label{eq:SSA}
  \end{equation}

  \noindent\textbf{Step-Localization Accuracy (SLA):}
  Pose accuracy is measured using normalized Maximum Symmetry-Aware Surface
  Distance (MSSD). Continuous local-\(Z\) symmetries are declared for all Nema17
  objects so that equivalent rotations are not penalized. For target-object
  diameter \(d\), we compute:
  \begin{equation}
  e_{target}
  =
  \frac{\mathrm{MSSD}_{target}}{d},
  \quad
  e_{assembly}
  =
  \frac{\mathrm{MSSD}_{assembly}}{d},
  \quad
  e_{step}
  =
  \frac{1}{2}(e_{target}+e_{assembly}).
  \label{eq:normalized_mssd}
  \end{equation}

  Step Acc@\({0.15d}\) is reported as the primary fixed-threshold score, together
  with accuracies at \(0.05d\), \(0.10d\), and \(0.20d\). Target AUC, Assembly
  AUC, and SLA-AUC summarize threshold accuracy over \(0\) to \(0.50d\), which is the main continuous pose metric.

\clearpage
\subsection{Quantitative results}
Table \ref{tab:quantitative_results} summarises the evaluation of PVRA and baselines on Nema17 test split. The number of samples evaluated for each method differ because each method can fail for different reasons: PVRA can fail during keypoint voting, CAD-ICP-PCA can fail due to insufficient segmented points for registration, and FoundationPose can reject small masks. These sample counts are therefore interpreted together with accuracy. 
\vspace{-2mm}
  \begin{table}[h!]
  \caption{Quantitative evaluation results for the Nema17 test
  split using symmetry-aware normalized MSSD. ICP-GT denotes CAD-ICP-PCA with
  annotation masks, ICP-PVRA denotes CAD-ICP-PCA with masks derived from PVRA
  segmentation, FP-GT denotes FoundationPose with annotation masks, and FP-PVRA
  denotes FoundationPose with masks derived from PVRA segmentation.}
  
  \label{tab:quantitative_results}
  \centering
  \footnotesize
  \renewcommand{\arraystretch}{1.25}
  \setlength{\tabcolsep}{3pt}
  \begin{tabular}{|p{0.30\textwidth}|c|c|c|c|c|}
  \hline
  \textbf{Metric / Method}&
  \textbf{PVRA} &
  \textbf{ICP-GT} &
  \textbf{ICP-PVRA} &
  \textbf{FP-GT} &
  \textbf{FP-PVRA} \\
  \hline
  Samples & 280 & 215 & 227 & 282 & 227 \\
  \hline
  SSA & 0.833 & -- & -- & -- & -- \\
  \hline
  Target AUC & 0.745 & 0.576 & 0.521 & 0.941 & 0.770 \\
  \hline
  Assembly AUC & 0.790 & 0.416 & 0.420 & 0.921 & 0.881 \\
  \hline
  SLA-AUC & 0.757 & 0.379 & 0.337 & 0.927 & 0.727 \\
  \hline
  Step Acc@0.15d & 0.893 & 0.344 & 0.286 & 0.989 & 0.771 \\
  \hline
  \end{tabular}
  \end{table}

\vspace{-1mm}
  \begin{figure}[!hb]
  \centering
  \includegraphics[width=0.72\textwidth]{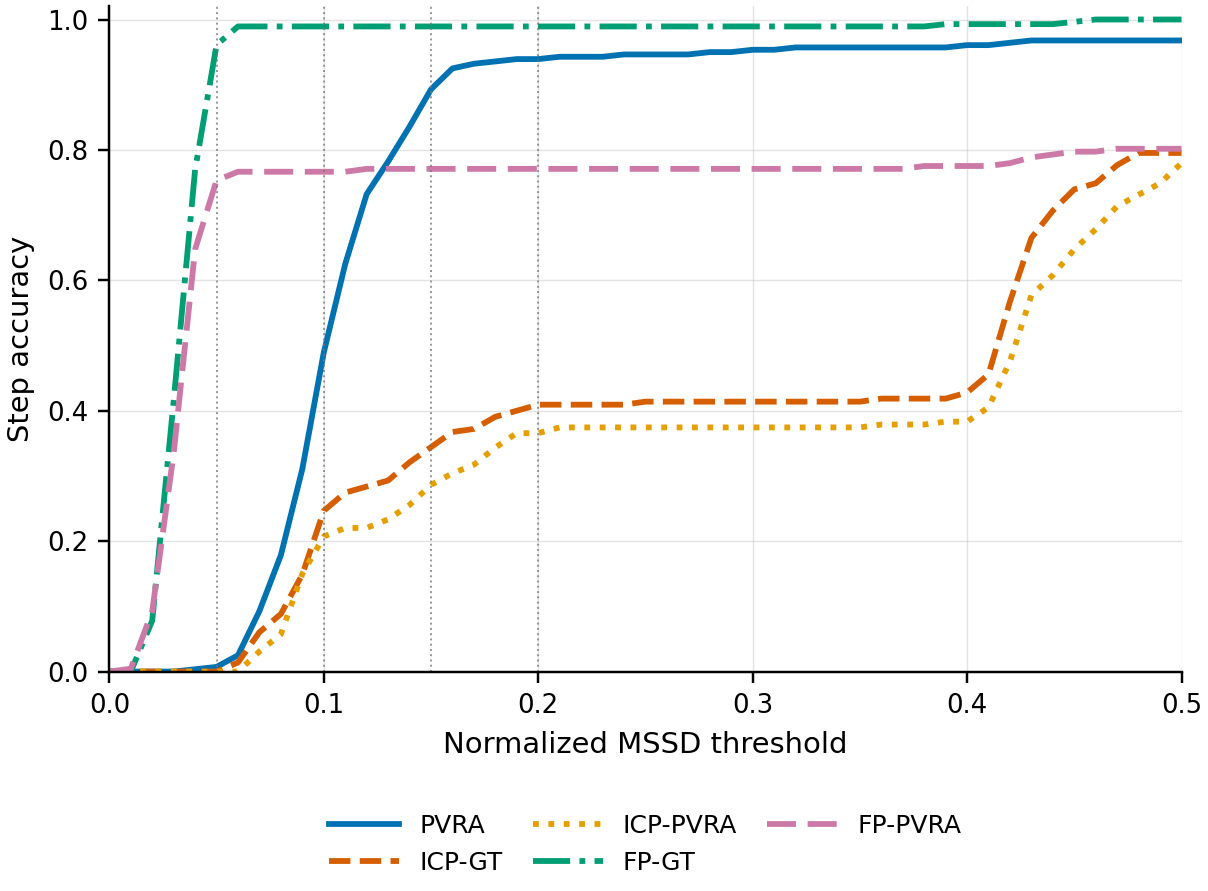}
  \caption{Step-level threshold accuracy curves.}
  \label{fig:threshold_curves}
  \end{figure}

Fig~\ref{fig:threshold_curves} shows a comparison of corresponding step-level threshold-accuracy curves. FoundationPose with ground-truth masks reaches high accuracy at very strict thresholds, while PVRA improves sharply between \(0.05d\) and \(0.15d\) ,while CAD-ICP-PCA curves fail to reach the accuracy of learning-based methods consistently. 

\subsection{Qualitative results}

\vspace{1mm}
\begin{center}
\newcommand{\qualw}{0.285\textwidth}
\newcommand{\qmetricfont}{\scriptsize}
\newcommand{\qmetrics}[3]{\parbox{\qualw}{\centering\qmetricfont T/A: #1 mm\\[-0.2mm]\(e_{\mathrm{step}}=#2\), #3}}
\setlength{\tabcolsep}{1.5pt}
\renewcommand{\arraystretch}{0.62}

\begin{tabular}{@{}p{0.085\textwidth}cc@{}}
 &
 \textbf{global-idx 32} &
 \textbf{global-idx 555} \\[-0.4mm]
\textbf{PVRA} &
\begin{tabular}{@{}c@{}}
\includegraphics[width=\qualw]{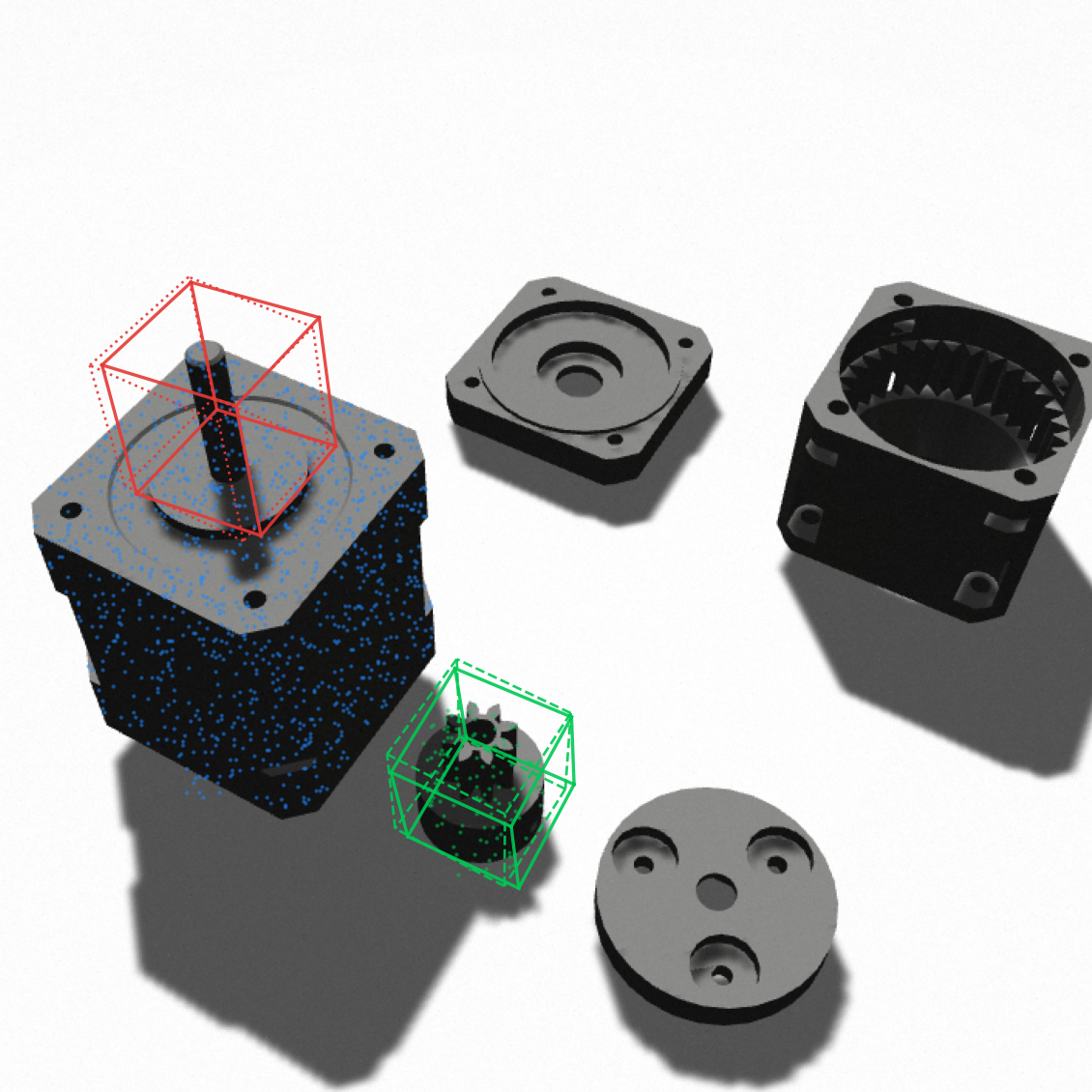}\\[-0.8mm]
\qmetrics{2.8/3.0}{0.082}{pass}
\end{tabular}
&
\begin{tabular}{@{}c@{}}
\includegraphics[width=\qualw]{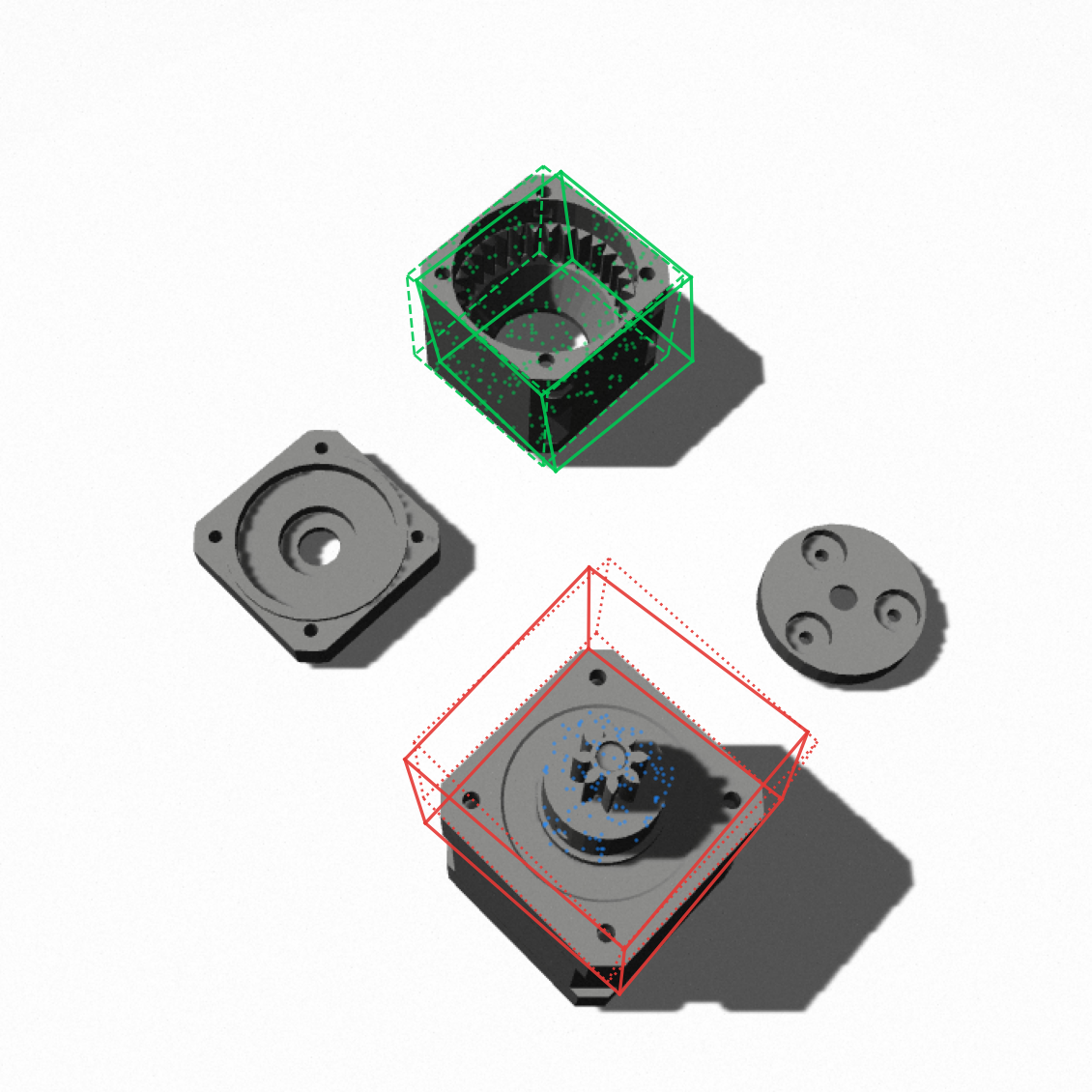}\\[-0.8mm]
\qmetrics{7.2/4.0}{0.084}{pass}
\end{tabular}
\\[1.0mm]

\textbf{FP} &
\begin{tabular}{@{}c@{}}
\includegraphics[width=\qualw]{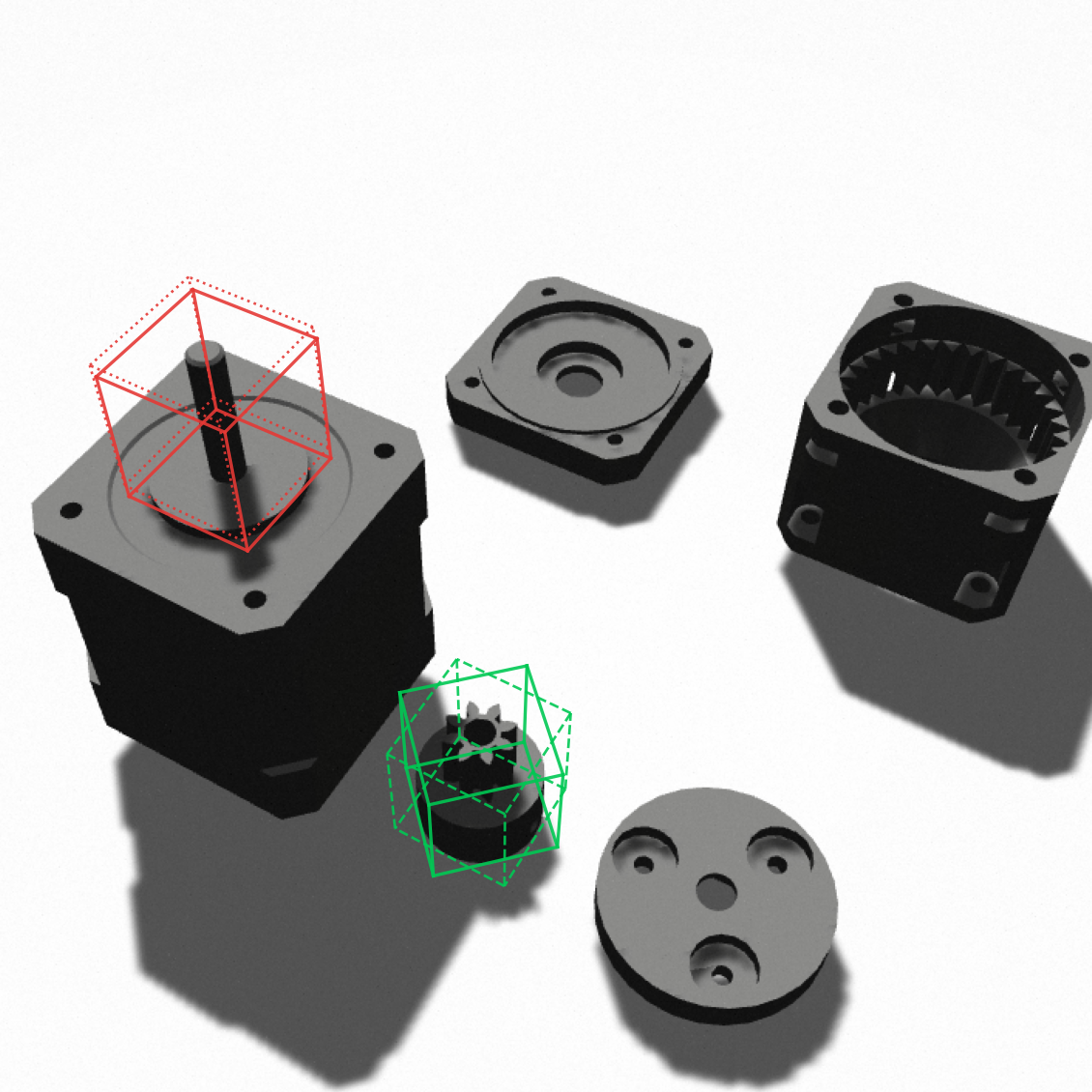}\\[-0.8mm]
\qmetrics{0.5/2.4}{0.041}{pass}
\end{tabular}
&
\begin{tabular}{@{}c@{}}
\includegraphics[width=\qualw]{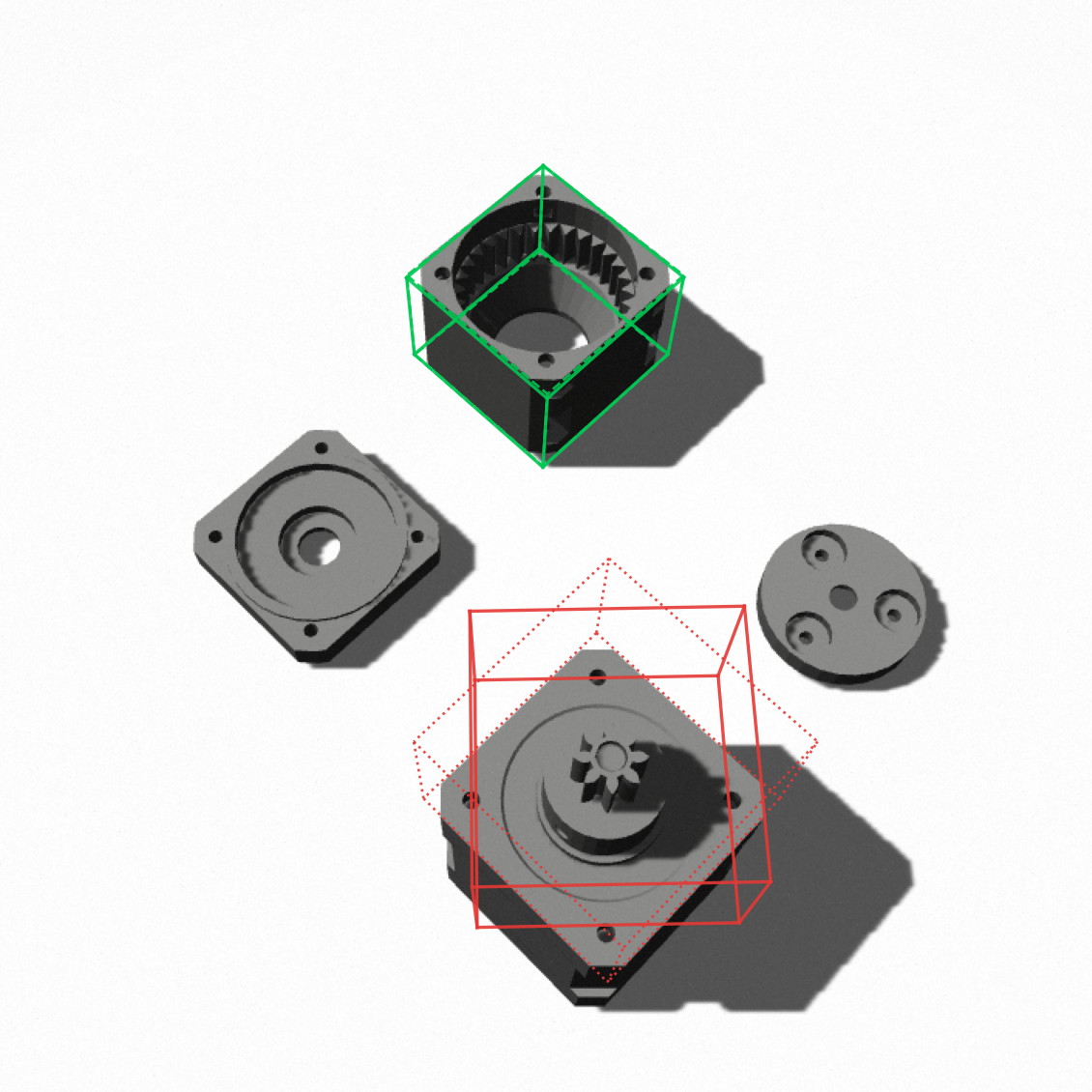}\\[-0.8mm]
\qmetrics{4.0/1.9}{0.044}{pass}
\end{tabular}
\\[1.0mm]

\textbf{ICP} &
\begin{tabular}{@{}c@{}}
\includegraphics[width=\qualw]{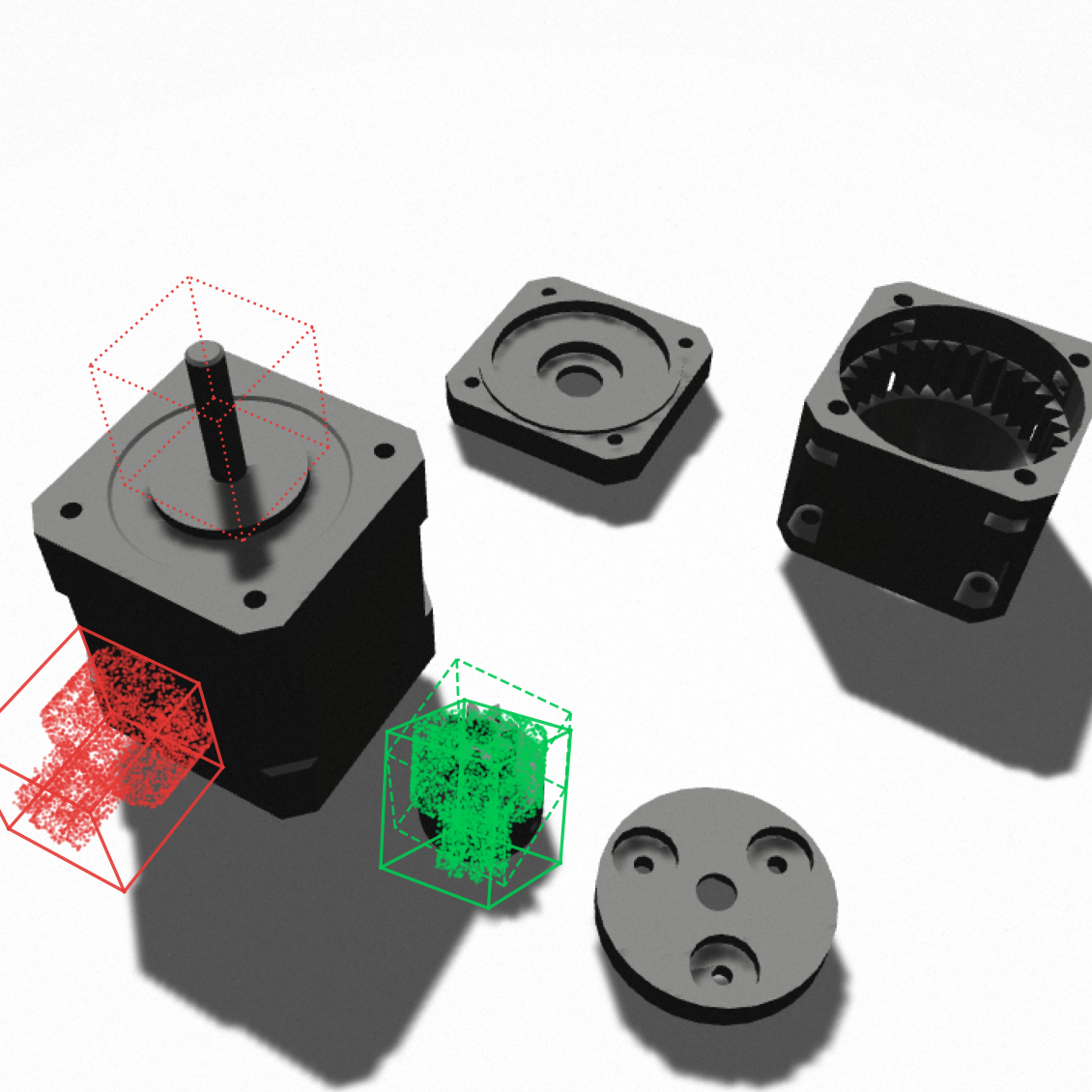}\\[-0.8mm]
\qmetrics{28.7/65.4}{1.314}{fail}
\end{tabular}
&
\begin{tabular}{@{}c@{}}
\includegraphics[width=\qualw]{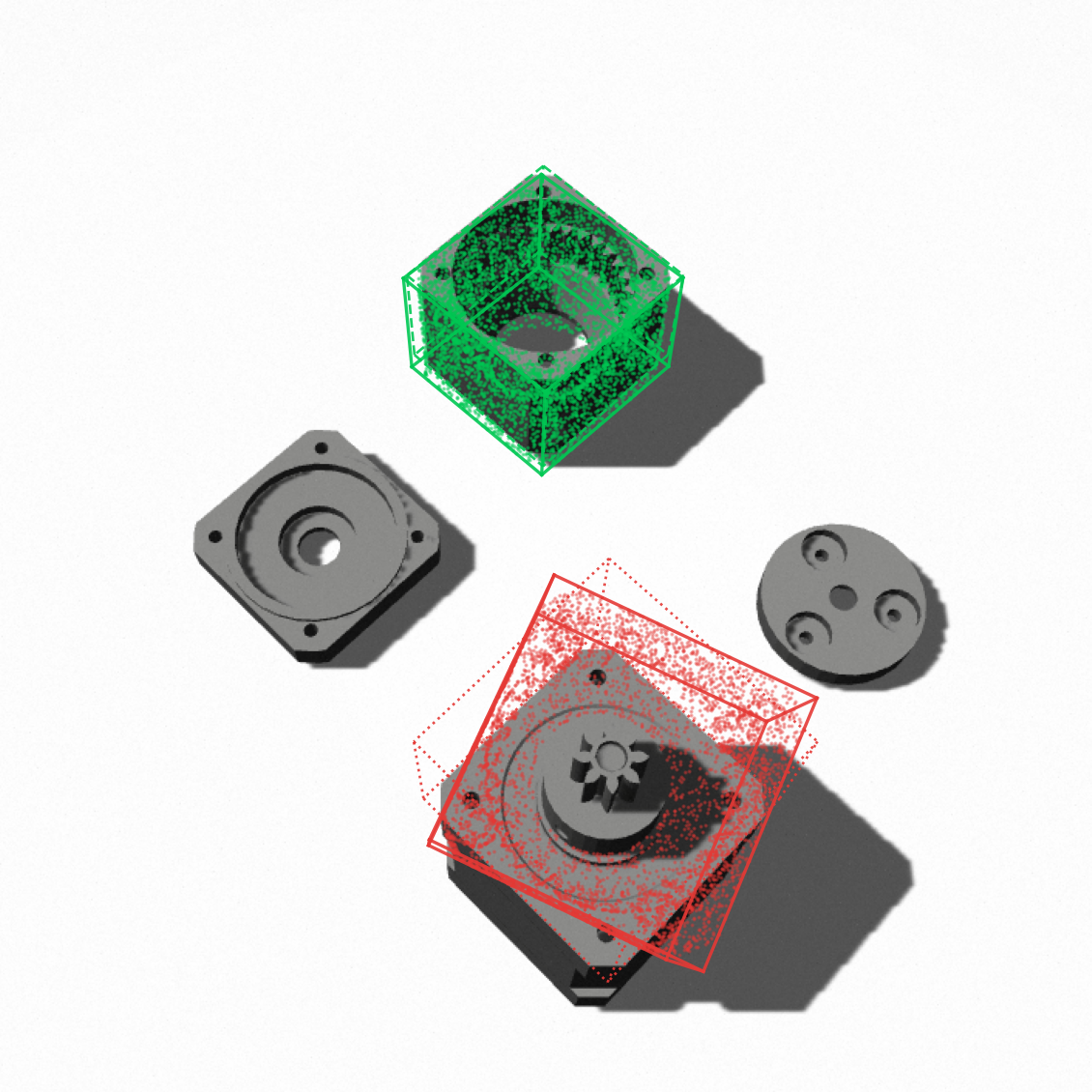}\\[-0.8mm]
\qmetrics{6.7/12.5}{0.143}{pass}
\end{tabular}
\end{tabular}
\vspace{-1mm}

\captionof{figure}{Qualitative comparison on two Nema17 test instances. Columns
show assembly instances, while rows compare PVRA, FoundationPose (FP), and
CAD-ICP-PCA (ICP). Green boxes indicate target-pose estimates and red boxes
indicate assembled-target estimates; dashed or dotted boxes indicate the
corresponding ground-truth poses. T/A reports target and assembled-target MSSD
in millimetres. \(e_{\mathrm{step}}\) is the normalized step MSSD, and pass/fail
indicates whether the sample satisfies Step Acc@\(0.15d\).}
\label{fig:qualitative_results}
\end{center}

\clearpage

\section{Discussion}\label{sec:discussion}

The evaluation reveals several methodological and empirical findings on assembly-aware pose prediction. First, CAD registration based methods are less effective when the scene consists of partial pointclouds. CAD-ICP-PCA performed poorly with both ground-truth masks and PVRA-predicted masks compared to other methods. The partial occlusions in progressive assembly make it further challenging for registration-based methods. Second, methods that highly dependent on accurate object masks are not reliable for assembly-aware pose prediction. FoundationPose predictions are highly accurate with ground-truth masks but accuracy drops with sparse PVRA-predicted masks while still performing well under strict thresholds compared to PVRA. However, its step SLA-AUC declines indicating that PVRA provides more consistent step-level pose accuracy among all threshold values. Third, FoundationPose with PVRA-predicted masks processed 53 fewer samples than PVRA. It  demonstrates robustness of PVRA under unavailability of dense object silhouettes as masks and uncertain or under-represent object boundaries. Fourth, FoundationPose baseline require prior knowledge on specific target and base object relationship corresponding to assembly sequence and relative pose transformation from base object to assembly pose. In contrast, PVRA predicts both poses from an RGB-D pointcloud representation in an end-to-end pipeline without explicit knowledge of the assembly instance while achieving a competitive accuracy.  

The evaluation uses a synthetic dataset of a progressive assembly. A real-world deployment may introduce additional challenges such as sensor noise, calibration errors, reflections and lighting conditions. While domain transferability remains a future research direction of this work, the simulation-based baseline comparison contributes an important reproducible benchmark for perception-based progressive assembly. As evident from the literature study, the availability of real-world structured datasets that represent progressive assemblies are limited. This is due to the complexity of capturing synchronized RGB-D frames and annotated 6-DoF poses in a real-world setting. Therefore, this controlled setting is essential for isolating the methodological question; can assembly context-aware model achieve comparable performance to existing object-centric pose estimation methods?. 

As a summary, the results show that PVRA addresses an important gap in robotic assembly manipulation that is not yet completely solved using object-centric perception. The average Nema17 object diameter is approximately 51 mm; therefore, the reported Step Acc@\(0.15d\) threshold corresponds to a normalized symmetry-aware surface-distance tolerance of about 7-8 mm for an average part. PVRA achieves a Step Acc@\(0.15d\) of 0.8929 which indicates 89.29\% of the total predictions are accurate under the threshold tolerance. This demonstrates the model's capability to learn spatial, relational and temporal dependencies within a progressive assembly and predict actionable outputs for downstream manipulation tasks. 


\clearpage
\section{Conclusion}\label{sec:conclusion}
This work presented a task-aware keypoint-based framework for progressive robotic assembly and a quantitative evaluation compared with existing object-centric perception methods. The evaluation confirms the ability of the model to learn assembly-task specific dependencies and produce accurate actionable outputs for robotic manipulation. Furthermore, this work motivates extending assembly-specific awareness towards broader task-specific awareness learning which can benefit a wider range of robotic manipulation applications beyond the scope of progressive assembly.

\begin{credits}
\subsubsection{\ackname} Project funding was received from - Helsinki Institute of Physics' Technology Programme (project; ROBOT) and Research Council of Finland under Grant no. 369003 (PERFORM). The authors wish to acknowledge CSC - IT Center for Science, Finland, for computational resources.
\end{credits}

\bibliographystyle{splncs04}
\bibliography{refs}

\end{document}